\documentclass[11pt,letterpaper]{article}
\usepackage[utf8]{inputenc}
\usepackage[margin=1in]{geometry}

\usepackage{graphicx}
\usepackage{array}
\PassOptionsToPackage{hyphens}{url}
\usepackage{url}
\IfFileExists{pdftexcmds.sty}{\usepackage{hyperref}}{\providecommand{\hypersetup}[1]{}}
\hypersetup{colorlinks=true,linkcolor=black,citecolor=black,urlcolor=blue,pdftitle={Monitoring Unplanned Construction Using Satellite Change Detection},pdfauthor={Harsh Kashyap and Taranum Wasu}}
\makeatletter
\newenvironment{inlinefigure}{\par\medskip\noindent\begin{minipage}{\textwidth}\def\@captype{figure}\centering}{\end{minipage}\par\medskip}
\makeatother
\title{Monitoring Unplanned Construction Using Satellite Change Detection}
\author{Harsh Kashyap \and Taranum Wasu}
\date{\small Thapar Institute of Engineering and Technology, Patiala, India\\hkashyap\_be19@thapar.edu \quad twasu\_be20@thapar.edu}
\begin{document}
\maketitle

\begin{abstract}
Cities need land for homes, shops, roads, and public services. New construction changes that land. It can harm natural areas and water bodies, add to waste, and strain public services. NITI Aayog's 2021 report stated that 65\% of 7,933 urban settlements in India had no master plan. That number tells us about planning coverage at the time. It does not tell us whether any single building was authorised.

Artificial intelligence, which lets computers learn patterns from data, can help find physical changes by comparing satellite images taken on different dates. We test simple methods on two public datasets from outside India. First, we turn each image into a set of numbers using a pretrained neural network, a model already trained to recognise patterns in images. We feed those numbers to standard machine-learning classifiers that predict land categories and physical changes. The methods reach about 95 percent accuracy at spotting built-up land and about 83 percent accuracy at saying whether a small pair of images shows change, where each pair gets one label. We also train a small neural network on a standard change-detection benchmark. Separately, we run a preliminary descriptive check on an Agra sample of 48 patches, 7 of them positive. The results show that satellite images carry information about physical change. They do not show whether construction was planned or permitted.

\end{abstract}
\section{Introduction}
Growing cities need more than new buildings. They need homes, workplaces, roads, and public services. They also have to protect land and natural resources. Construction can shrink natural areas, harm water bodies, create demolition waste, and put pressure on public infrastructure. Some of this damage is hard to undo once land is built on. Planning rules help cities manage growth. To apply these rules, authorities need reliable information about changes on the ground. Several European jurisdictions have pursued policies to limit land development for decades, but automated monitoring remains a separate challenge. France's Climate and Resilience Law, for example, aims for zero net land artificialisation by 2050. It also sets an intermediate goal of halving the consumption of natural, agricultural, and forest land over 2021--2031 \cite{r22}. These targets illustrate the need to track physical changes when making land-use decisions.

India faces a large planning gap. NITI Aayog's 2021 report stated that 65\% of India's 7,933 urban settlements had no master plan \cite{r19}. That is a dated measure of planning coverage. It is not a count of illegal buildings, and a missing master plan does not make any particular building illegal.

Construction also creates waste. In February 2024, the Ministry of Housing and Urban Affairs cited estimates of 150--500 million tonnes of construction and demolition waste each year in India. It also estimated that India would need to add 700--900 million square metres of commercial and residential space each year by 2030 \cite{r20}. The waste range is wide because it is an estimate, not a national count, and it does not cover unplanned construction alone. The space figure describes future building demand, not future waste. India's Environment (Construction and Demolition) Waste Management Rules, 2025 took effect on 1 April 2026. They bring in extended producer responsibility, recycling targets, and online compliance monitoring \cite{r21}. Together, these planning and waste concerns raise the need for reliable information about changes to land and buildings.

Satellite images can supply some of that information. Images of the same place taken on different dates can show where visible features have changed. Automated methods can help authorities, planners, researchers, and local communities find those changes and look at them more closely. The goal is not to automate enforcement. It is to make physical changes easier to find and review.

This paper examines three questions. Can we identify new built-up areas by comparing satellite images from two dates? Do these methods work across Indian cities? Can predictions with higher confidence reduce errors as an early step towards human review? Our completed experiments address the first question, using two public datasets from outside India. We add a small preliminary descriptive check on an Agra sample. We have not tested performance across Indian cities, so the second question is still open. For the third question, we look at confidence-based selection on a benchmark change-detection task. We have not tested a process where people review the uncertain predictions.

One distinction shapes the whole study: physical change versus legal status. Words like ``unplanned'' and ``unauthorised'' describe legal or administrative conditions. A satellite image does not show them by itself. A built-up-area classifier identifies land cover, and a change detector finds differences between two images. Neither can tell us whether construction follows planning rules without more information, such as zoning boundaries, permits, or master plans.

The paper reports three completed contributions and describes two planned extensions. The first contribution evaluates frozen image features with standard classifiers, repeats experiments across random seeds, and compares random data splits with splits that keep patches from the same source scene together. The second examines classification thresholds, abstention, calibration, and learning curves for a patch-level change task. Abstention means leaving a prediction unanswered rather than making an automatic decision. The third trains and evaluates a compact Siamese change-detection network on LEVIR-CD using a CPU. A Siamese network processes two images through connected network branches so that their features can be compared. We plan two extensions: testing stronger change-detection models on Indian data, and building a confidence-based process for human review. Neither is done yet. The results here do not show performance across India, transfer across Indian cities, or detection of unauthorised construction.

\section{Related Work}
Our review follows the main stages of the study. We first examine methods for classifying land cover in satellite images. We then review datasets and models for detecting changes between two dates. Finally, we discuss resources for Indian imagery and research that connects image-based change detection with planning status.

\subsection*{Land-cover classification from satellites}
Land-cover classification assigns image areas to categories such as buildings, vegetation, and water. Researchers have studied this task extensively. EuroSAT provides 64 $\times$ 64 Sentinel-2 image patches in ten classes and is widely used as a benchmark \cite{r1}. The DeepGlobe 2018 challenge extended this research to larger datasets covering multiple regions and attracted broad research participation \cite{r2}.

Other studies have mapped buildings across large areas. Sirko et al. detected buildings across continents using high-resolution satellite images and released the resulting building footprints publicly \cite{r3}. Ji et al. released an open dataset of aerial and satellite images, along with models for extracting buildings from them \cite{r4}. These studies show that satellite images can support built-up-area mapping when suitable labels are available. Our approach uses a simpler method. Rather than training a dedicated land-cover network, we extract image features from a general-purpose network and use them to train standard classifiers.

\subsection*{Change detection: datasets}
Change detection compares images of the same place taken at different times. When the images come from two dates, the task is called bi-temporal change detection. LEVIR-CD contains 637 high-resolution image pairs showing building changes, along with masks that mark changed pixels. It is the main change-detection benchmark used in this study \cite{r5}. OSCD provides smaller pairs of multi-spectral Sentinel-2 images for detecting urban change \cite{r6}. SECOND goes beyond binary change detection by labelling what each side of an image pair contains \cite{r7}. The xBD dataset labels building damage across disaster events and shows how change detection can support relief work \cite{r8}. These datasets provide labels for land cover, physical change, or building damage. They do not provide the planning or permit labels needed to determine whether construction was authorised. Neither of the two main datasets in our experiments is Indian, and neither identifies whether a change was planned.

\subsection*{Change detection: models}
Early deep-learning methods used two connected network branches to process images taken at different times. These models are called Siamese networks. The fully convolutional Siamese designs developed by Caye Daudt et al. remain an important reference for this approach \cite{r9}. Later models improved change detection in different ways. Transformer-based models such as BIT \cite{r10} and ChangeFormer \cite{r11} use attention, a method that helps a model determine which parts of its image features are most relevant to one another. Both report strong results on LEVIR-CD. Smaller models such as TinyCD \cite{r12} and SNUNet-CD \cite{r13} aim to maintain good accuracy while using less computing power.

Two broad reviews examine this field. Shi et al. review change detection methods based on artificial intelligence (AI) and their open challenges \cite{r14}. Khelifi and Mignotte compare methods and datasets through a meta-analysis \cite{r15}.

Our Siamese model sits at the lightweight end of this field. We are not trying to match larger models. We want to see what a small model can do when training is limited to a CPU.

\subsection*{Resources for India}
Public labelled datasets for Indian cities remain limited. IndiaSat provides pixel-level land-cover labels for imagery from Landsat-7, Landsat-8, and Sentinel-2 for 2019 \cite{r16}. ICCD provides 1024 $\times$ 1024 image pairs with building-change masks for ten Indian cities. The images were collected between 2020 and 2024, and the dataset was released with a semi-supervised baseline \cite{r17}. A semi-supervised method learns from labelled examples alongside data that do not have labels.

Both datasets give us material for studying Indian imagery, but their access conditions differ and we have not obtained all the data needed for evaluation. Section 4 describes the data we could inspect and the experiments still to be done.

\subsection*{From imagery to planning status}
Detecting physical change and deciding whether it was unplanned are different tasks. The second requires information beyond the image itself, and fewer studies address it directly.

A 2026 systematic review examines deep-learning and geospatial-AI methods for detecting unauthorised construction using satellite and UAV imagery \cite{r18}. Geospatial AI applies artificial intelligence to information linked to specific places. UAV refers to an unmanned aerial vehicle, such as a drone. The accepted manuscript describes how cadastral information, which records land parcels and their boundaries, and zoning maps can serve as reference data. Detected structures are compared with official plans before legal compliance is assessed.

This distinction sets the limits of our study. Satellite images can show physical change, but administrative records are needed to judge whether a change follows planning rules. The review does not validate our proposed workflow, and it does not give city-specific permit labels for India. As far as we know, no public benchmark links building changes in Indian cities to planning or permit outcomes. Our planned work aims to fill that gap.

\section{Methods}
Our experiments use three main ideas: frozen image features, classical classifiers, and repeated random seeds. Together, they provide a straightforward way to test whether satellite images contain useful information about land cover and physical change. Frozen features are numerical descriptions of an image produced by a neural network that has already been trained on a large collection of general images. We keep the network's learned parameters unchanged. Each image passes through the network, and the resulting numbers become the input to another model.

Classical classifiers are standard machine-learning models that learn to assign labels from these numerical features. We use logistic regression and random forests. Logistic regression estimates the probability of a class. A random forest combines predictions from many decision trees, which make decisions through a series of feature-based rules. Each forest in our experiments contains 400 trees. We also repeat experiments with different random seeds. A seed determines the random choices used to create a data split. We use five seeds for these experiments. A single split may produce an unusually good or poor result, so repeated runs help show how much results depend on that choice. We report the average score and standard deviation, which measures variation between runs.

\subsection{Feasibility experiments}
\textbf{Built-up classification.} We resize each image to 112 $\times$ 112 pixels and pass it through a frozen ImageNet MobileNetV3-Small network. We use the PyTorch/torchvision implementation with IMAGENET1K\_V1 weights. We replace the final classifier with an identity layer so that the network returns its 576-dimensional feature vector directly. The network runs in evaluation mode, which keeps its learned parameters fixed. We resize RGB images to 112$\times$112, scale their pixel values to [0, 1], and normalise them using ImageNet means (0.485, 0.456, 0.406) and standard deviations (0.229, 0.224, 0.225). We train either logistic regression or a random forest with 400 trees on these features. Both classifiers use class-balanced weights, which give additional weight to less common classes during training. The dataset contains up to 800 images from each original class, giving 8,000 images in total, including 2,400 built-up images.

\textbf{Change classification.} To compare images from two dates, we calculate the absolute difference between their image features. We also calculate four measurements of the pixel differences: the mean, standard deviation, maximum, and fraction of values above 0.2. Together, these measurements describe differences in image features and pixel values. A patch receives a positive label if more than 2\% of its change mask is foreground. This threshold defines the task used in our experiment. It is not a validated minimum for meaningful change. The model assigns one label to the entire image pair rather than a separate label to each pixel. We therefore treat this as patch-level classification, not segmentation, which identifies changed areas pixel by pixel. We use a random forest with 400 trees.

\textbf{Data splits.} For random splits, we reserve 30\% of the samples for testing while preserving the proportions of the labels. We generate five splits using seeds 0--4. All models use the same split for each seed.

We also use a scene-grouped split for LEVIR-CD. Instead of assigning individual patches to the test set, we hold out 30\% of the source scenes. All patches from the same scene stay together because they may be closely related. This reduces the risk that closely related patches appear in both training and testing. Unlike the random split, the grouped split does not preserve label proportions, so the shares of positive and negative examples may differ in the test set. For each of seeds 0--4, we hold out 39 of 128 scenes, containing 624 patches. The positive counts are 229, 223, 210, 202 and 217. We set the fixed-feature classifier settings in advance. Logistic regression uses up to 2000 iterations, and the random forest uses 400 trees. Both use class-balanced weights, and the features are standardised so that they are on a comparable scale. We do not tune these fixed-feature classifiers or use a validation set for their settings. The separate calibration and SiamCD validation procedures are described later. The held-out samples do not influence the choice of model settings.

\textbf{Features and metrics.} We calculate the four pixel-difference measurements using the two 112 $\times$ 112 RGB images after scaling their pixel values to [0, 1]. Each measurement uses all pixels and channels of the absolute difference.

We report accuracy, balanced accuracy, F1, and AUROC. Accuracy is the share of predictions that are correct. Balanced accuracy gives equal weight to performance on the two classes. F1 combines precision and recall. Precision is the share of predicted positives that are correct, while recall is the share of actual positives that are found. The positive class is built-up land for EuroSAT and changed patches for LEVIR-CD. AUROC measures whether positive examples tend to receive higher scores than negative examples. It considers different classification thresholds rather than relying on a single cutoff. A score of 0.5 represents chance-level ranking. For the main experiments, we classify a sample as positive when its predicted probability is p $\geq$ 0.5. The historical sensitivity and selective-accuracy script instead uses $p>0.5$, as explained in Section 5.

\subsection{Planned pipeline (not run)}
The next stage will use the public reference implementations of BIT and TinyCD. We plan to train them on LEVIR-CD using their published default settings. We will then evaluate them on ICCD without additional training and after fine-tuning on Indian data.

We also plan to estimate uncertainty using a deep ensemble. This method combines predictions from multiple trained models to estimate how certain their predictions are. Using validation data, we would select an abstention threshold that meets a target rate of missed changes. Abstention means leaving some predictions unanswered rather than making an automatic decision.

For segmentation, our main error measure will count changed pixels that the model incorrectly labels as unchanged. We divide this count by the total number of ground-truth changed pixels in the evaluated city. Pixels set aside for review will not count as automatic misses. We will report the area sent for review separately.

The proposed operating rule would select the highest automatic coverage for which the validation missed- change fraction is at most 5\%. Coverage is the share of cases receiving an automatic prediction. We would then report the missed-change fraction and review-area fraction on the test data without changing the threshold. The 5\% bound is a provisional research target, not an authority-approved acceptable loss. We would also report sensitivity at 1\%, 5\% and 10\%, along with recall for each city.

We will measure missed buildings separately at the object level. This means counting buildings that were missed rather than counting only their pixels. A pixel-based limit alone could hide cases where an entire small building is missed. None of these planned experiments has been run.

\section{Datasets}
Table 1 summarises the datasets used or considered in this study. Figures 1, 2 and 6 show dataset examples.

\textbf{EuroSAT.} EuroSAT contains Sentinel-2 land-use patches in 10 classes \cite{r1}. We load the images through torchvision. We define a binary task in which built-up land includes \{Residential, Industrial, Highway\}. The other seven classes form the negative class. Because the images come from Europe, this dataset serves only as a proxy for built-up land classification.

\textbf{LEVIR-CD.} LEVIR-CD contains image pairs from before and after a change, along with masks that identify changed pixels \cite{r5}. Our experiments use a subset of 2,048 patches at 256 $\times$ 256 pixels. The files come from the test shard of the Hugging Face copy ericyu/LEVIRCD\_Cropped256. We identify 128 source scenes by parsing the filenames. We did not verify the geographical region covered by the patches.

We created our own data splits from this subset, as described in Section 3. The held-out samples evaluate our experiments. They are not an official benchmark test set.

\begin{table}[!htbp]\centering\footnotesize\setlength{\tabcolsep}{3pt}
\caption{Datasets. Counts for EuroSAT and LEVIR-CD describe the subsets used in the experiments, calculated from the files used in the runs. ICCD figures describe only the public sample.}\label{tab:1}
\begin{tabular}{p{0.21\textwidth}p{0.08\textwidth}p{0.17\textwidth}p{0.17\textwidth}p{0.12\textwidth}p{0.15\textwidth}}
\hline
Dataset & Region & Unit & Evaluation size & Positives & Status \\
\hline
EuroSAT & Europe & 64 $\times$ 64 patch & 8,000 & 2,400 (30\%) & used \\
LEVIR-CD (evaluation subset) & non-Indian & 256 $\times$ 256 pair & 2,048 (128 scenes) & 731 (35.7\%) & used \\
ICCD sample (Agra) & India & 1024 $\times$ 1024 pair & 3 labelled pairs & 7 of 48 patches & preliminary check \\
ICCD (full) & India (10 cities) & 1024 $\times$ 1024 pair & not obtained & -- & access pending \\
IndiaSat & India & pixel / point & not obtained & -- & export pending \\
\hline\end{tabular}
\end{table}
\textbf{ICCD.} ICCD contains labelled and unlabelled image pairs from ten Indian cities. It is hosted on IEEE DataPort (DOI 10.21227/0ppm-8374; full archive 6.47--7.36 GB). Access to the full archive requires an IEEE DataPort subscription. A free IEEE account does not provide access to the files.

We inspected only the public 19-file Agra sample. On 11 October 2026, we sent the dataset submitter a private message through IEEE DataPort to request access. No reply was recorded in October 2026. We could not find licence text in the GitHub repository, the DataPort page, or the README. We must resolve access, terms of use, and per-city counts before running further experiments.

The Agra sample supports only a small preliminary descriptive check. It contains 48 patches, including 7 positive patches. This sample cannot establish model accuracy, generalisation across Indian cities, or the ability to identify unauthorised construction.

\textbf{IndiaSat.} IndiaSat provides labelled land-cover points and imagery obtained through Google Earth Engine scripts \cite{r16}. The GitHub training shapefile we inspected contains 1,873 labelled polygons: 94 greenery, 112 water, 1,391 built-up and 276 barren-land polygons.

These are polygon counts, not counts of independent pixel observations. They also differ from the publication's 1,855 polygons and 180,414 extracted pixels. We therefore cannot assume that the repository version exactly reproduces the published sample.

The author scripts require authenticated Earth Engine access and the authors' assets. Both users/chahatresearch/IndiaSat and users/chahatdel/traindata\_14\_May returned ``Collection asset not found''. We did not export any imagery.

The repository shapefile could be uploaded as a replacement once the sample mismatch, export region, and sensor protocol have been resolved.

\section{Experiments}
\subsection{Feasibility experiments}
Table 2 reports the mean and standard deviation across five random seeds. A random seed determines the data split and other random choices made during an experiment. Table 3 reports the results for each seed separately. Together, the tables show how much the results change when different samples are used for training and testing.

We first ran the original authors' scripts on our machine using a single seed. On EuroSAT, logistic regression, a model that estimates class probabilities from input features, achieved an accuracy of 0.948 and an AUROC of 0.985. Accuracy is the proportion of predictions that are correct. The area under the receiver operating characteristic curve (AUROC) measures how well a model ranks positive examples above negative examples. An AUROC of 1.0 indicates perfect ranking, while 0.5 is the expected value for random ranking.

On LEVIR-CD, the same evaluation approach achieved an accuracy of 0.837 and an AUROC of 0.923. Both values match the seed-0 results in Table 3 and fall within the ranges observed across all five seeds. This agreement provides an initial check that the original scripts run correctly on our machine.

\begin{figure}[!htbp]\centering
\includegraphics[width=0.98\textwidth,height=0.70\textheight,keepaspectratio]{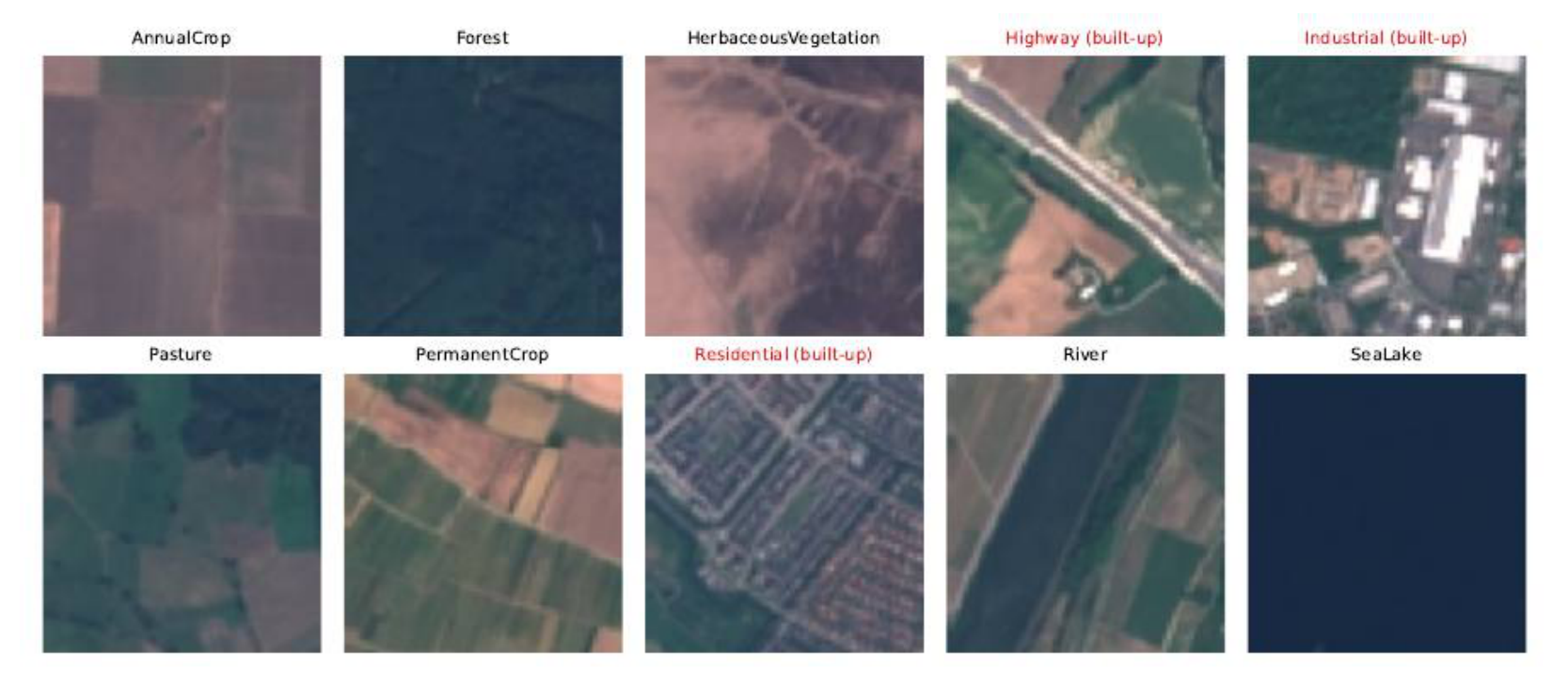}
\caption{EuroSAT: one example image from each of the ten classes, in sorted file order. Red titles identify the three classes used to define the built-up positive class. These land-cover labels do not identify the exact locations of buildings within an image. Dataset source: Helber et al. \cite{r1}, copyright 2023 Patrick Helber, MIT License; contains Copernicus Sentinel data. The dataset license notice is included with the source package.}\label{fig:1}
\end{figure}
\begin{table}[!htbp]\centering\footnotesize\setlength{\tabcolsep}{3pt}
\caption{Feasibility experiments, mean and standard deviation over five split seeds. Random splits assign 70\% of samples to training and 30\% to testing. Grouped LEVIR-CD splits hold out 30\% of scenes and keep all patches of each scene together; patch counts and class proportions need not match the random splits. The standard deviation measures variation across splits, not uncertainty across cities. The positive class is built-up land for EuroSAT or more than 2\% changed pixels for LEVIR-CD. All data are non-Indian. Source: results/multiseed.json.}\label{tab:2}
\begin{tabular}{p{0.16\textwidth}p{0.18\textwidth}p{0.14\textwidth}p{0.14\textwidth}p{0.14\textwidth}p{0.14\textwidth}}
\hline
Task & Split / model & Acc & Bal. acc & F1 & AUROC \\
\hline
EuroSAT built-up & random, LogReg & 0.947 $\pm$ 0.002 & 0.940 $\pm$ 0.003 & 0.912 $\pm$ 0.003 & 0.984 $\pm$ 0.002 \\
EuroSAT built-up & random, RF & 0.930 $\pm$ 0.006 & 0.897 $\pm$ 0.008 & 0.875 $\pm$ 0.012 & 0.982 $\pm$ 0.002 \\
LEVIR-CD proxy & random, RF & 0.835 $\pm$ 0.011 & 0.792 $\pm$ 0.016 & 0.735 $\pm$ 0.023 & 0.916 $\pm$ 0.013 \\
LEVIR-CD proxy & scene-grouped, RF & 0.827 $\pm$ 0.023 & 0.776 $\pm$ 0.031 & 0.707 $\pm$ 0.049 & 0.899 $\pm$ 0.029 \\
\hline\end{tabular}
\end{table}
\begin{figure}[!htbp]\centering
\includegraphics[width=0.98\textwidth,height=0.70\textheight,keepaspectratio]{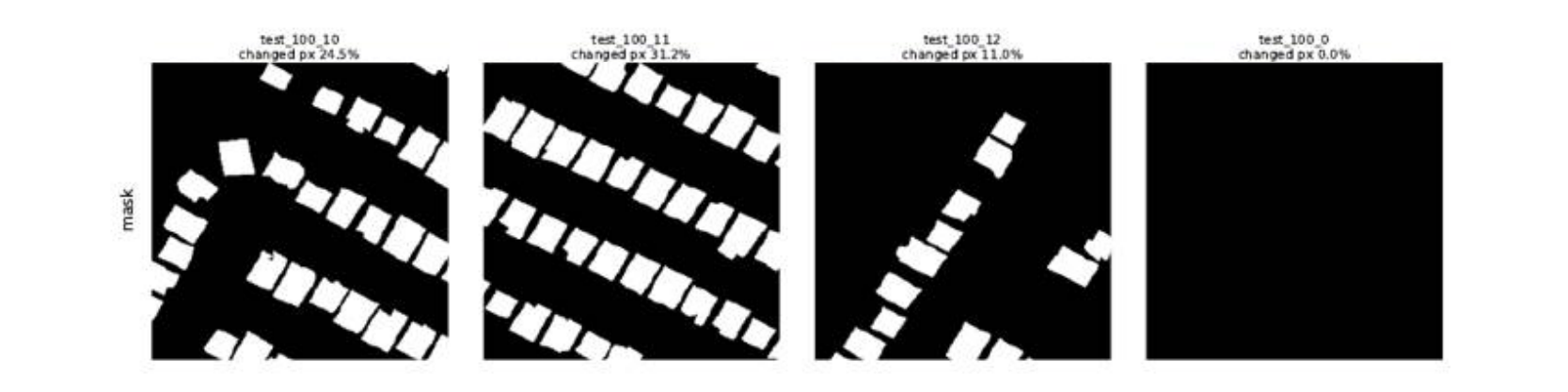}
\caption{Ground-truth masks from four LEVIR-CD evaluation-subset patches. The photographic panels are omitted. The first three patches have more than 5\% changed pixels; the last has at most 2\%, in sorted file order. These examples were not selected based on model predictions. Under the 2\% rule, the last patch is negative. Source: Chen and Shi \cite{r5}; dataset annotations shown for academic use.}\label{fig:2}
\end{figure}
\begin{table}[!htbp]\centering\footnotesize\setlength{\tabcolsep}{3pt}
\caption{Accuracy / AUROC for each seed. The same script was rerun with per-seed output. The means reproduce Table 2.}\label{tab:3}
\begin{tabular}{p{0.23\textwidth}p{0.134\textwidth}p{0.134\textwidth}p{0.134\textwidth}p{0.134\textwidth}p{0.134\textwidth}}
\hline
Task / model & Seed 0 & Seed 1 & Seed 2 & Seed 3 & Seed 4 \\
\hline
EuroSAT LogReg & 0.948/0.985 & 0.944/0.982 & 0.945/0.982 & 0.949/0.986 & 0.948/0.985 \\
EuroSAT RF & 0.931/0.983 & 0.918/0.978 & 0.937/0.984 & 0.934/0.982 & 0.930/0.981 \\
LEVIR-CD random & 0.837/0.923 & 0.839/0.924 & 0.850/0.929 & 0.828/0.910 & 0.818/0.893 \\
LEVIR-CD scene-grouped & 0.833/0.917 & 0.784/0.849 & 0.841/0.906 & 0.825/0.888 & 0.851/0.935 \\
\hline\end{tabular}
\end{table}
Both datasets support the planned classification experiments. The results also raise a question about how training and test data should be separated. The next section examines whether keeping patches from the same LEVIR-CD scene together changes the measured performance.

\subsection{Effect of grouping by scene}
We compared random splits with scene-grouped splits across five seeds. In a random split, each patch is assigned to the training or test set independently. In a scene-grouped split, all patches from the same scene remain in the same set. This prevents patches from one scene from appearing in both training and testing. It also reduces the chance that a model benefits from similarities between related images across the two sets.

For LEVIR-CD, mean AUROC fell from 0.916 $\pm$ 0.013 with random splits to 0.899 $\pm$ 0.029 with scene-grouped splits. We also calculated the difference between the two methods for each seed by subtracting the grouped-split AUROC from the random-split AUROC. The mean paired difference was 0.017 $\pm$ 0.038. The five differences were 0.005, 0.075, 0.023, 0.022 and $-$0.042.

The results did not show a consistent pattern across the five seeds. Scene grouping produced a higher AUROC for one seed, and variation between seeds was large relative to the average difference. These five runs were not enough to determine whether grouping consistently changed performance or whether the observed gap reflected differences between splits. We therefore extended the comparison to 30 seeds, numbered 0--29, using the same code.

Across the 30 runs, mean AUROC was 0.913 $\pm$ 0.011 for random splits and 0.903 $\pm$ 0.018 for scene-grouped splits. The mean paired difference was 0.010 $\pm$ 0.022. Random splits produced a higher AUROC in 20 of the 30 seeds (Figure 5). The average difference remained small relative to the variation across splits.

These additional runs provide a broader comparison of the two splitting methods on LEVIR-CD. They do not, however, represent 30 independent datasets. Every run uses the same 2,048 patches and changes only the division between training and testing. The results describe variation across repeated splits of one dataset, not variation across independent cities or datasets. We do not report a significance test because the repeated splits do not provide independent observations.

One possible explanation for the higher average AUROC under random splits is that related patches can appear in both training and testing. A model may perform better on a test image when it has already encountered similar images during training. The results are consistent with this explanation, but they do not establish that it caused the difference.

We could not run the same scene-grouping comparison for EuroSAT because scene identifiers were not available to us. Nearby tiles or near-duplicate images could make its results more optimistic, but we have not measured this effect.

\begin{figure}[!htbp]\centering
\includegraphics[width=0.98\textwidth,height=0.70\textheight,keepaspectratio]{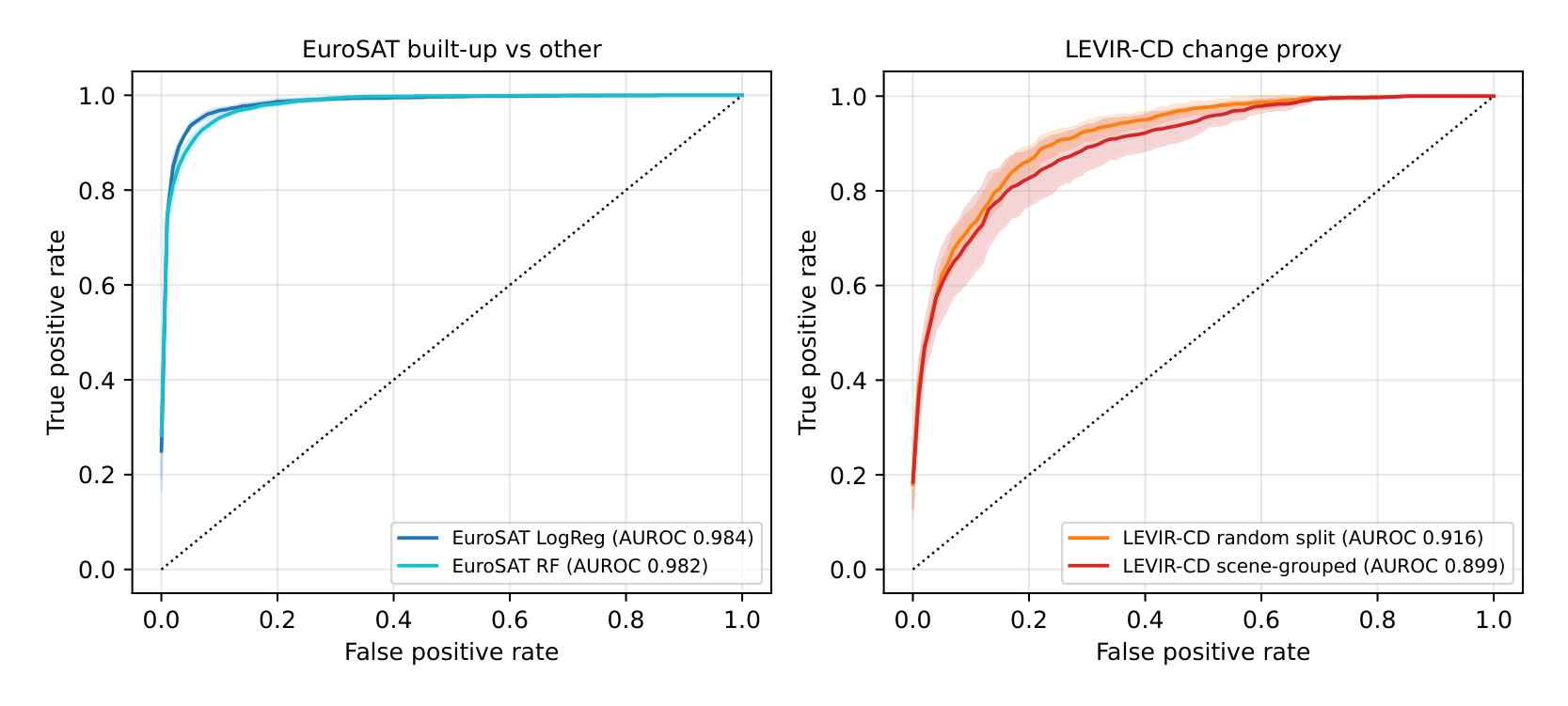}
\caption{ROC curves on the held-out samples, averaged over five seeds. Random splits hold out 30\% of patches, while grouped splits hold out 30\% of scenes. The shaded band spans one standard deviation of the true positive rate at each false positive rate. Left: EuroSAT built-up task. Right: LEVIR-CD change classification.}\label{fig:3}
\end{figure}
\begin{figure}[!htbp]\centering
\includegraphics[width=0.98\textwidth,height=0.70\textheight,keepaspectratio]{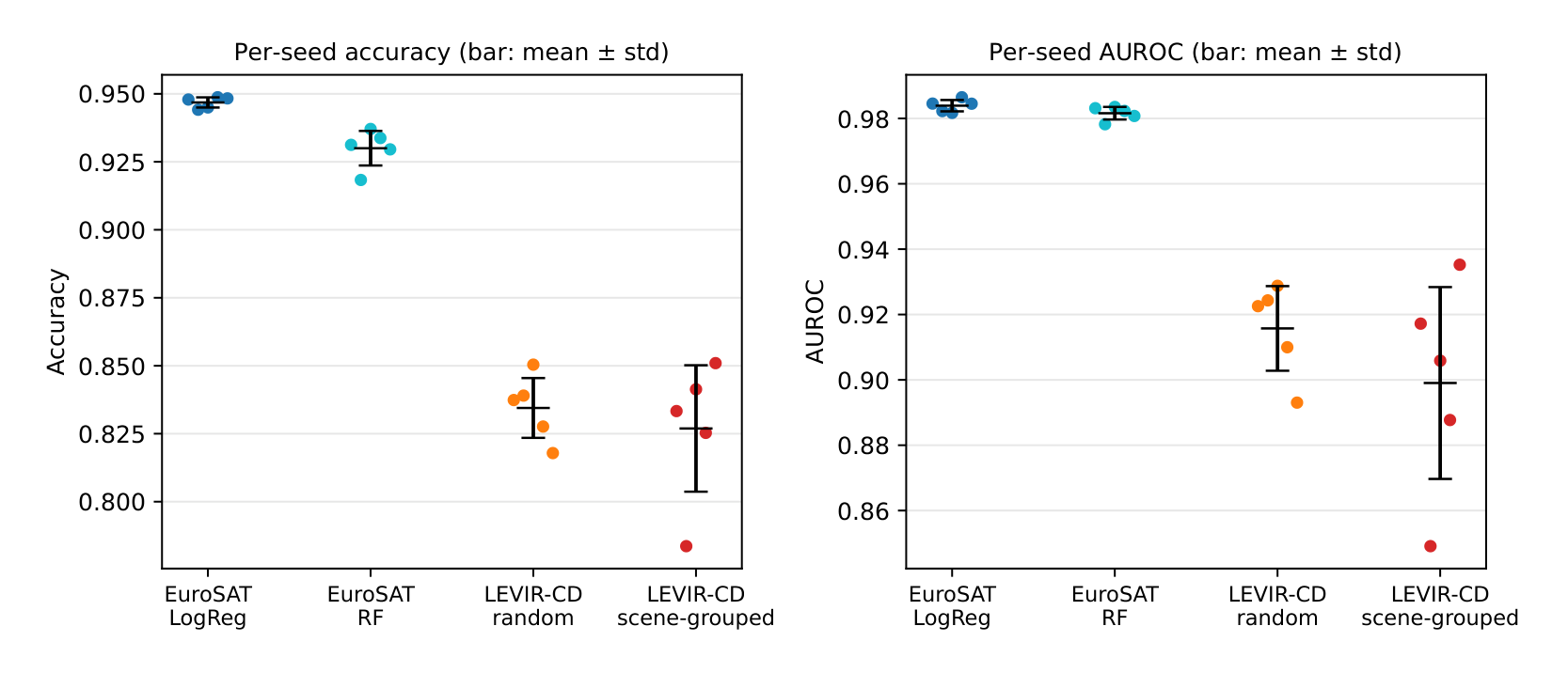}
\caption{Accuracy on the left and AUROC on the right for each task and seed. Black bars show the mean and one standard deviation. The LEVIR-CD scene-grouped runs vary most between seeds.}\label{fig:4}
\end{figure}
\begin{figure}[!htbp]\centering
\includegraphics[width=0.72\textwidth,height=0.70\textheight,keepaspectratio]{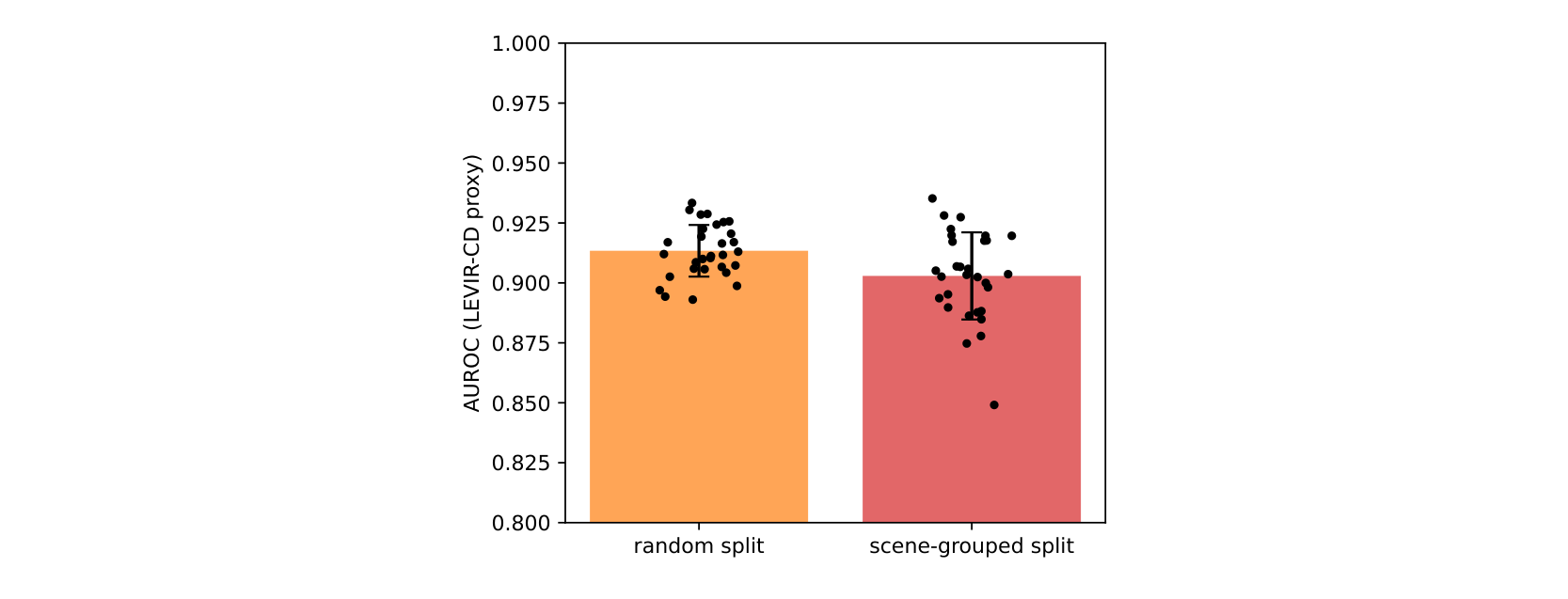}
\caption{LEVIR-CD AUROC under random and scene-grouped splits over 30 seeds. Bars show the mean and one standard deviation. Dots show individual seeds. The five-seed values in Table 2 are seeds 0--4 of these runs.}\label{fig:5}
\end{figure}
\subsection{Indian-imagery preliminary check (sample only)}
We applied the random forest trained on LEVIR-CD features to 48 patches from the Agra sample. We created these patches by dividing the three labelled Agra image pairs in Figure 6 into non-overlapping 4 $\times$ 4 crops. Each crop measured 256 $\times$ 256 pixels. As in the main change-classification experiment, we labelled a patch positive if more than 2\% of its pixels had changed and negative otherwise.

The model achieved an accuracy of 0.854 and an AUROC of 0.737. It predicted 6 of the 48 patches as positive, representing 12.5\% of the sample. The ground truth contained 7 positive patches.

These results describe performance on three labelled image pairs from Agra. The sample contains only 48 patches and 7 positive examples. It is too small to establish how well the model performs across Indian cities. The scores are preliminary descriptive observations, not evidence of general transfer to Indian imagery.

\begin{figure}[!htbp]\centering
\includegraphics[width=0.98\textwidth,height=0.70\textheight,keepaspectratio]{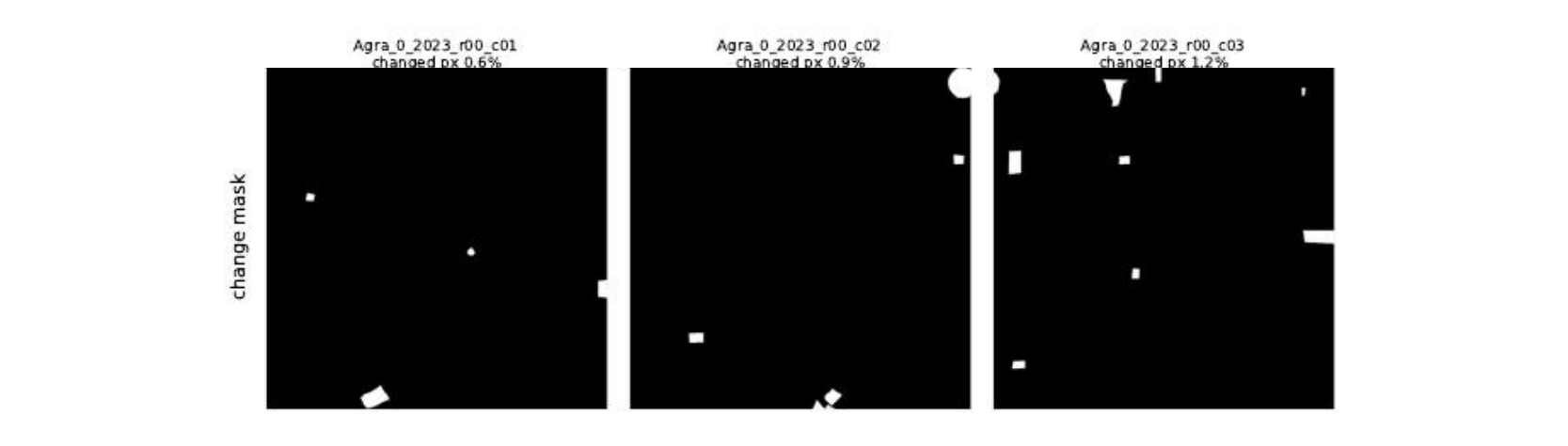}
\caption{Ground-truth change masks for the three labelled ICCD public Agra sample pairs from 2022 and 2023. The photographic panels are omitted. Changed pixels account for 0.6\%, 0.9\% and 1.2\% of each pair. Each whole pair falls below the 2\% threshold in Section 3. Positive patches in the preliminary check arise only after dividing each pair into $4\times4$ crops. Dataset and annotations: Srivastava et al. \cite{r17}, IEEE DataPort, doi:10.21227/0ppm-8374, Creative Commons Attribution; cropped to show annotations only.}\label{fig:6}
\end{figure}
\subsection{Sensitivity, abstention and learning-curve checks}
Before examining the additional experiments, we resolved a small difference between two reported accuracy summaries. Full-coverage accuracy was 0.826 $\pm$ 0.024 in the additional analysis and 0.827 $\pm$ 0.023 in Table 2. The difference came from how the scripts handled predictions with a probability of exactly 0.5. The extra.py script predicted positive only when $p>0.5$, whereas multiseed\_eval.py predicted positive when p $\geq$ 0.5.

We reran the evaluation using the original cached features, labels and scene identifiers. The rerun reproduced both summaries under their respective prediction rules. For seeds 0--4, the numbers of predictions with exactly $p=0.5$ were 2, 1, 1, 4 and 0. We retain the reported values because each follows its script's prediction rule. The difference is small, and the two rules explain why the summaries are not identical.

We then ran four additional analyses using the cached LEVIR-CD features, the scene-grouped splits for seeds 0--4 and the same random forest. Figure 8 summarises the results.

\textbf{Label threshold.} We changed the percentage of changed pixels required for a patch to count as positive. A higher threshold changes which patches belong to each class. Results at different thresholds therefore describe different classification tasks, rather than the same task under different model settings. As the threshold increased from 0.5\% to 10\%, the positive rate fell from 42.0\% to 19.8\%.

AUROC increased from 0.892 $\pm$ 0.028 at 0.5\%, through 0.899 $\pm$ 0.029 at 2\%, to 0.924 $\pm$ 0.021 at 10\%. One possible explanation is that patches with larger changed areas are easier to distinguish from patches with less change. The results show that AUROC increased with the threshold, but they do not establish why.

\textbf{Abstention.} We ranked test patches by the random forest's confidence and kept predictions only for the most confident patches. Coverage is the proportion of patches for which the model makes a prediction. At full coverage, accuracy was 0.826 $\pm$ 0.024. At 50\% coverage, accuracy rose to 0.940 $\pm$ 0.026. At 10\% coverage, it reached 0.997 $\pm$ 0.006.

The model was more accurate on the patches it selected when it answered only its most confident cases. However, we did not report the class composition of these selected patches. High accuracy alone does not show whether the model detects most changed patches. We also did not measure the workload for human reviewers. These results therefore do not establish whether confidence-based selection would support an effective human-review process.

\textbf{Learning curve.} We trained the model using 10\%, 25\%, 50\% and 100\% of the training split. The corresponding AUROC values were 0.865 $\pm$ 0.036, 0.884 $\pm$ 0.035, 0.891 $\pm$ 0.036 and 0.899 $\pm$ 0.031.

AUROC increased as the amount of training data grew. This pattern appeared across all four training sizes. The results do not establish how certain the gains between training sizes are.

\textbf{Calibration.} Calibration measures how closely a model's predicted probabilities match the observed frequency of positive outcomes. For example, among cases assigned a probability of 0.8, a well-calibrated model should produce positive outcomes in about 80\% of cases. We divided each training set by scene, using 75\% to fit the model and 25\% to calibrate its probabilities. We then evaluated the model on the same held-out scenes.

Before calibration, the random forest had an expected calibration error (ECE) of 0.112 $\pm$ 0.031 and a Brier score of 0.138 $\pm$ 0.014. ECE measures the difference between predicted confidence and observed outcome frequency across groups of predictions. The Brier score measures the average squared difference between predicted probabilities and actual binary outcomes. We calculated ECE using 10 equal-width probability bins.

We then applied isotonic recalibration, a method that adjusts predicted probabilities to better match observed outcomes. ECE fell to 0.068 $\pm$ 0.020, and the Brier score fell to 0.128 $\pm$ 0.021. AUROC changed from 0.896 $\pm$ 0.033 to 0.890 $\pm$ 0.032. This small change suggests that the model's ranking of examples remained broadly similar even though its probability estimates changed.

The calibration experiment used only 75\% of the training data to fit the forest. This explains why its AUROC values differ slightly from those in Table 2. Figure 7 shows the reliability curves, which compare predicted probabilities with observed outcome frequencies. Calibration remained imperfect. These results do not establish that the model's confidence scores are reliable enough to guide review decisions on Indian data.

\begin{figure}[!htbp]\centering
\includegraphics[width=0.72\textwidth,height=0.70\textheight,keepaspectratio]{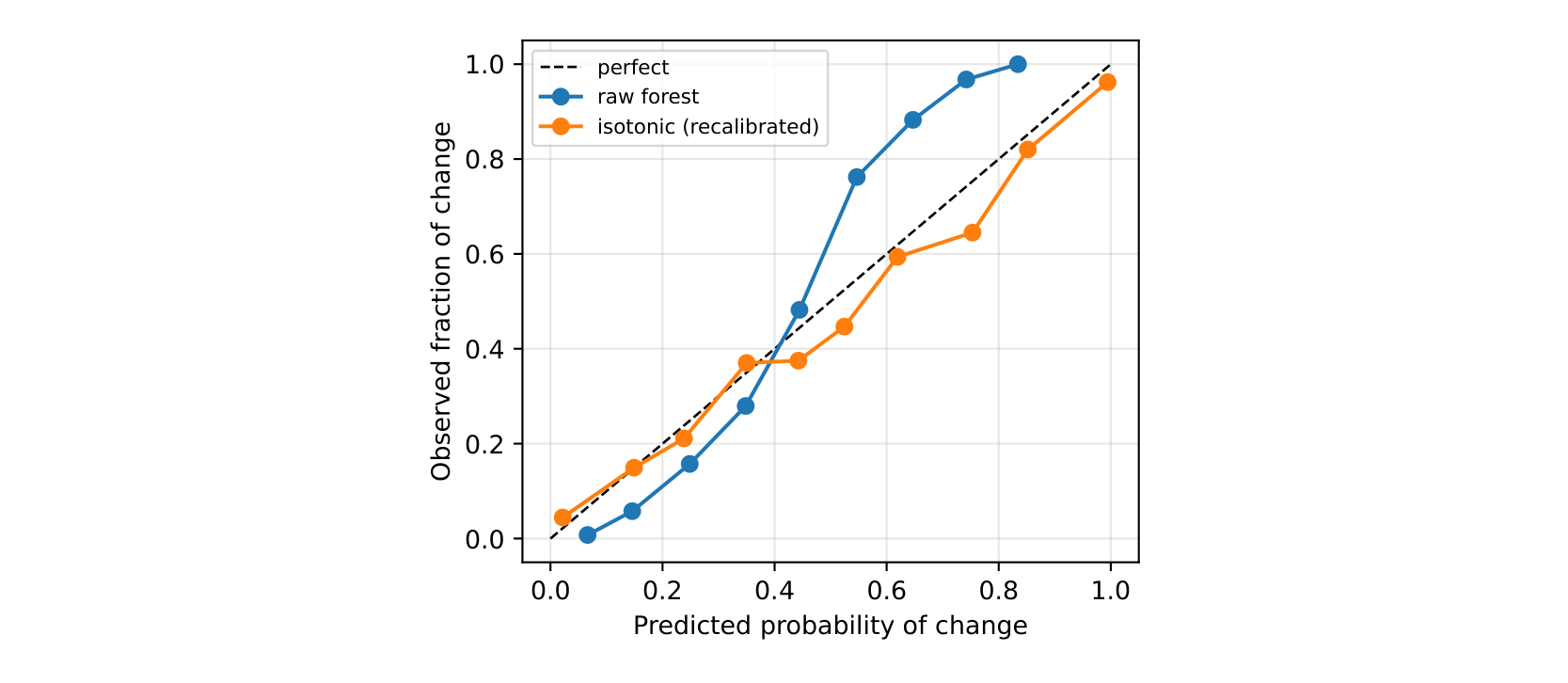}
\caption{Reliability diagram for the LEVIR-CD task under scene-grouped splits, pooled over held-out patches from seeds 0--4. Bins with fewer than 10 patches are omitted. The dashed line shows perfect calibration.}\label{fig:7}
\end{figure}
\begin{figure}[!htbp]\centering
\includegraphics[width=0.98\textwidth,height=0.70\textheight,keepaspectratio]{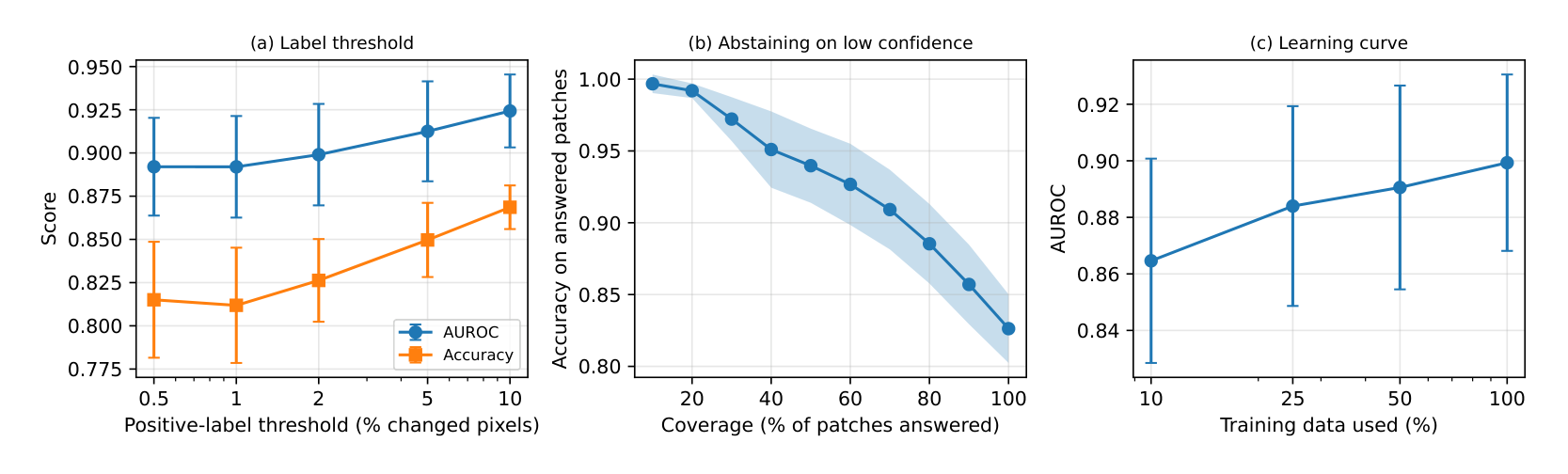}
\caption{LEVIR-CD results for scene-grouped splits, seeds 0--4, showing the mean and one standard deviation. Panel (a) shows AUROC and accuracy against the changed-pixel threshold used to label a patch as positive. Panel (b) shows accuracy among the most confident fraction of patches. High accuracy does not establish recall of changed patches. Panel (c) shows AUROC against the share of the training split used.}\label{fig:8}
\end{figure}
\subsection{A trained pixel-level change detector}
The earlier experiments classify whole image pairs using fixed features extracted from the images. They do not identify individual changed pixels. To test whether we could also locate changes, we trained a model on LEVIR-CD to predict whether each pixel had changed.

Our model, SiamCD, is a compact Siamese encoder. It processes both images using the same network weights, then compares their features at three scales using absolute differences. A lightweight U-Net-style decoder converts these comparisons into a one-channel change map. Each output value represents the predicted probability that a pixel has changed. The model has only 119 K parameters.

We trained SiamCD on 1,500 LEVIR-CD training pairs and 400 validation pairs at 128$\times$128 resolution. We used binary cross-entropy loss, which measures the difference between predicted probabilities and the true binary labels. Changed pixels are less common than unchanged pixels, so we set the positive-class weight to 10 to account for this imbalance. We used Adam to update the model's parameters and selected the checkpoint with the best validation IoU.

IoU, or intersection over union, measures how much the predicted change area overlaps with the ground-truth change area. It divides the area shared by both masks by the total area covered by either mask. Training ran on a CPU within a fixed 25-minute budget.

We evaluated SiamCD at pixel level on the full official LEVIR-CD test split, which contains 2,048 pairs.

SiamCD achieved a test IoU of 0.605, an F1 score of 0.754, precision of 0.726 and recall of 0.785. Precision measures the proportion of predicted changed pixels that were actually changed. Recall measures the proportion of actual changed pixels that the model detected. Table 5 reports the results, and Figure 11 shows examples of its predictions.

The comparison table in the TinyCD paper reports LEVIR-CD F1 scores of 0.8931 for BIT, 0.9040 for ChangeFormer and 0.9105 for TinyCD \cite{r12}. These are published results, not reruns under our subset-trained, CPU-only setup, and do not form a matched comparison. SiamCD is a small baseline trained on a subset of the data using a CPU. We did not attempt to match the published results.

This experiment shows that a pixel-level segmentation model can be trained and evaluated within our computing limits. It adds a pixel-level result to the patch-level classification results reported earlier. However, SiamCD was trained and tested on the same non-Indian benchmark. Its results do not establish whether the model transfers to Indian imagery.

\begin{table}[!htbp]\centering\footnotesize\setlength{\tabcolsep}{3pt}
\caption{Evidence boundaries. ``Tested'' does not mean validated for deployment.}\label{tab:4}
\begin{tabular}{p{0.17\textwidth}p{0.42\textwidth}p{0.31\textwidth}}
\hline
Question / objective & Completed evidence & Not established \\
\hline
RQ (i): physical change & EuroSAT land cover; LEVIR-CD patch classification; descriptive Agra sample; reported within-LEVIR-CD SiamCD segmentation baseline & Indian-city transfer or deployment quality \\
RQ (ii): transfer & Agra: 48 crops from 3 image pairs; no cross-city evaluation & Generalisation across the ten ICCD cities \\
RQ (iii): confidence & LEVIR-CD accuracy-coverage and calibration checks & Human-review quality or a validated operational threshold \\
Planning status & No planning or permit labels & Whether a detected change is unplanned or unauthorised \\
\hline\end{tabular}
\end{table}
\begin{figure}[!htbp]\centering
\includegraphics[width=0.98\textwidth,height=0.70\textheight,keepaspectratio]{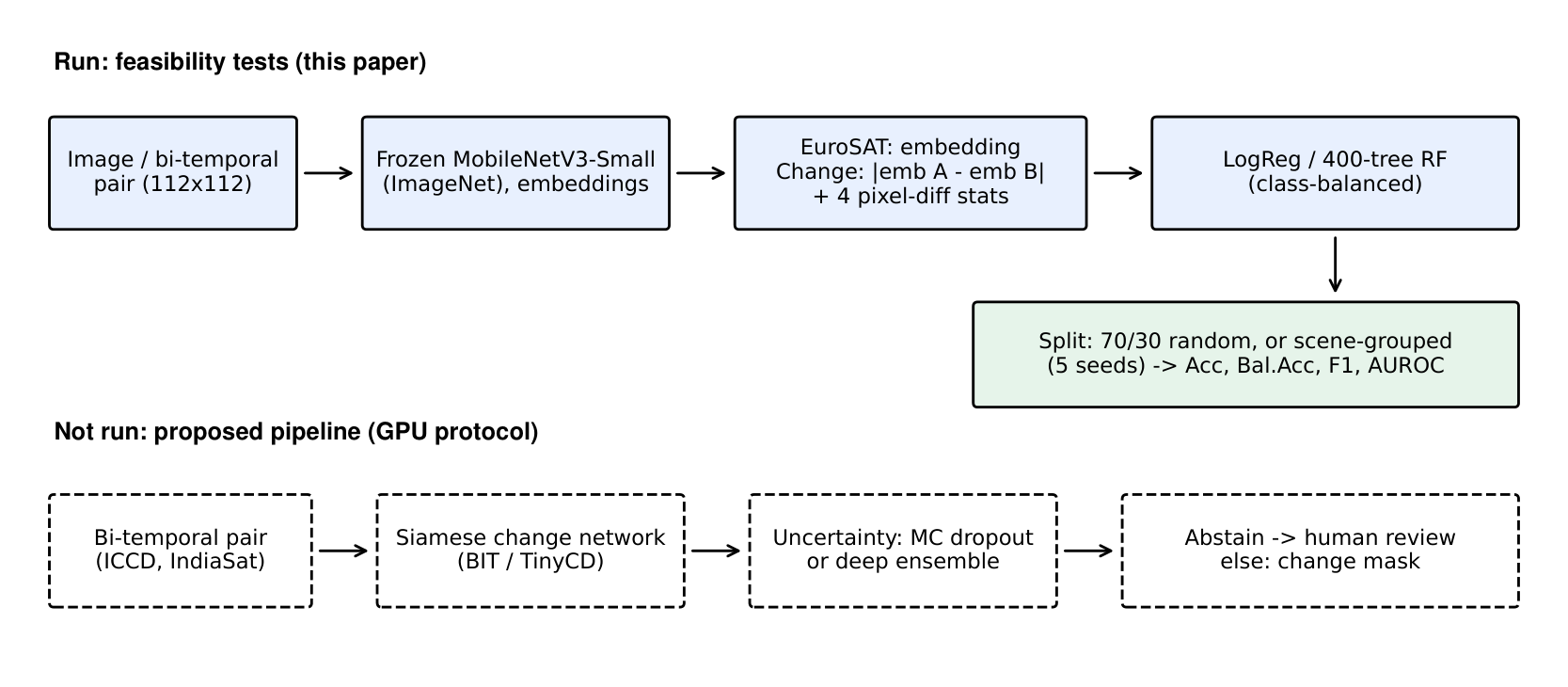}
\caption{Experimental and planned workflows. The upper row shows the experiments reported in this paper. We extract fixed image features and train standard classifiers, repeating the experiments with five different data splits. The lower row shows the workflow planned for future work, in which uncertain cases would be set aside for human review. Only the upper row has been run.}\label{fig:9}
\end{figure}
\begin{table}[!htbp]\centering\footnotesize\setlength{\tabcolsep}{3pt}
\caption{Trained change detector (SiamCD) on the full LEVIR-CD test split (2,048 pairs), at pixel level. This is a single CPU run. Published models provide context, not a matched comparison; see the comparison in Section 5.5. Source: results/cd\_baseline\_results.json.}\label{tab:5}
\begin{tabular}{lllll}
\hline
Model & IoU & F1 & Precision & Recall \\
\hline
SiamCD (119 K params, CPU, 1.5 k train) & 0.605 & 0.754 & 0.726 & 0.785 \\
\hline\end{tabular}
\end{table}
\section{Discussion}
The experiments show that fixed image features can distinguish built-up land from other land-cover classes in EuroSAT. They can also distinguish LEVIR-CD patches with more than 2\% changed pixels from patches below that threshold. Under the splits used in this study, both tasks achieve AUROC values well above 0.5, the level expected from random ranking.

The pixel-level SiamCD experiment provides a separate result. It shows that a small network can be trained and evaluated on LEVIR-CD using a CPU. Its test F1 score of 0.754 provides a baseline for this benchmark, but it is not a matched comparison with the published models discussed in Section 5.5.

These findings answer only part of our research questions. The experiments test physical-change tasks on the datasets we had. They do not show how the methods perform across Indian cities. The Agra check is still a small preliminary descriptive exercise with 48 patches and 7 positives. The confidence-based experiments measure accuracy among the patches we keep, not the quality of a human-review process. And none of our experiments uses planning or permit labels.

The next step is to test the methods on Indian data and compare candidate changes with planning records that match the image dates. We need that separate check before these results can support any decision about planning status.

\subsection*{Limitations}
The two main datasets use non-Indian imagery. The Agra sample is too small to establish performance across Indian cities. We use fixed ImageNet features at 112 px, one architecture and standard classifiers. The study is therefore a feasibility study, not a competitive evaluation of change-detection models.

The change-classification task uses a patch-level proxy with an arbitrary 2\% threshold, so its scores cannot be compared directly with pixel-level segmentation benchmarks. We use no planning, permit, or legal labels, so the study does not detect unplanned construction.

The EuroSAT split is not grouped by scene. Across 30 overlapping LEVIR-CD splits, the difference between random and scene-grouped AUROC is about one point. This estimate is approximate because all splits reuse the same dataset. The fixed-feature classifiers use held-out samples from a subset that we split ourselves, not an official benchmark test set. The separate SiamCD evaluation is reported on the official test split. The ICCD DataPort landing page lists Creative Commons Attribution, but the source-imagery redistribution terms and required credits still need checking. IndiaSat imagery redistribution terms also remain unverified.

\section{Conclusion}
Fixed image features can distinguish built-up land-cover labels and patch-level change labels on two non-Indian datasets under the splits used in this study. A small Siamese detector trained on a CPU also achieves a baseline score on the LEVIR-CD benchmark.

These results support further evaluation on Indian data. They do not show transfer to Indian cities, full-scale segmentation performance, or detection of unplanned construction. The next steps are to get ICCD access, run GPU-based baselines, and obtain planning or permit records that match the image dates. Those records are what a workflow needs before authorities can review candidate changes and decide their planning status.

\subsection*{Future work}
\begin{inlinefigure}
\includegraphics[width=0.98\textwidth,height=0.70\textheight,keepaspectratio]{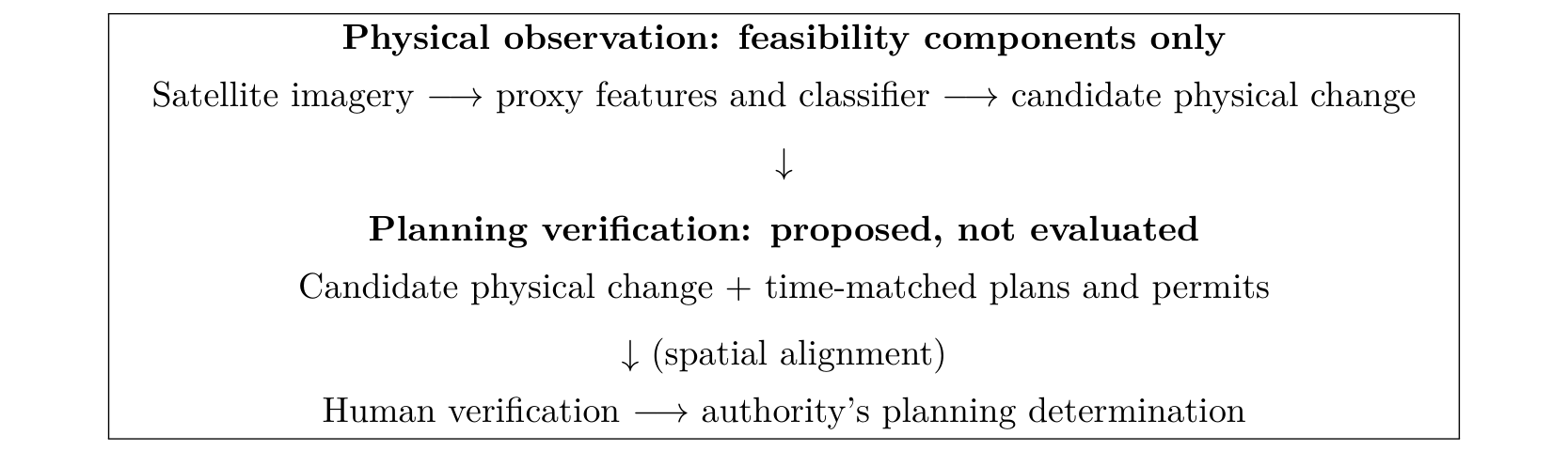}
\caption{The path from satellite imagery to a planning determination. The upper part shows the physical-change experiments reported here, using non-Indian benchmarks and the descriptive Agra sample. The lower part shows the proposed planning-verification process, which has not been evaluated. Physical change alone is not proof of a planning violation.}\label{fig:10}
\end{inlinefigure}
The remaining experiments require a GPU or datasets that we do not yet have. We have completed the within- LEVIR detector baseline on a CPU, achieving a test IoU of 0.605 and an F1 score of 0.754 on the full LEVIR-CD test split.

Training the larger reference models, BIT and TinyCD, to their published performance needs a GPU. We checked free computing options, but Colab and Kaggle both asked us to sign in before we could keep editing notebooks. We have postponed these experiments until we have GPU access. The official TinyCD repository includes a training script and a way to evaluate pretrained models, but results from pretrained models cannot replace a new training run.

We plan to test cross-city generalisation by leaving out each of the ten ICCD cities in turn. First, we will evaluate models trained on LEVIR-CD without further training on ICCD. Next, we will fine-tune the models on the other nine cities and test them on the city left out. We will report IoU and F1 for each city, along with the mean and standard deviation over at least three seeds. After examining the baseline results, we will choose a semi-supervised method for the unlabelled ICCD pairs. These experiments will replace the descriptive Agra check in Section 5 as the basis for studying cross-city performance.

For built-up land-cover classification in India, we plan to train and evaluate models on IndiaSat using spatially blocked cross-validation. This method keeps nearby areas in the same data split, reducing the chance that closely related locations appear in both training and test sets. Once the Earth Engine imagery export succeeds, we will compare these models with the EuroSAT-pretrained classifier.

We have already calculated calibration error and a risk--coverage curve for the fixed-feature classification task on LEVIR-CD. A risk--coverage curve shows how prediction errors change as the system answers more or fewer cases automatically. We still need to repeat the uncertainty and abstention checks for BIT and TinyCD. These tests will include the fraction of area sent for review at a fixed missed-change rate on ICCD. We have also completed the LEVIR-CD threshold check. Once ICCD access is available, we will repeat the threshold, abstention and learning-curve checks on that dataset.

Finally, identifying unplanned construction requires evidence beyond satellite imagery. Possible sources include building changes outside a published planned layout or master-plan boundary, and permit records from a local authority.

The Agra Development Authority is a possible starting point because our descriptive sample comes from Agra. Its official plan pages list Master Plan 2021, Master Plan 2031 and zonal-plan material (\url{https://www.adaagra.org.in/master-plan,} \url{https://www.adaagra.org.in/plan}). However, the existence of these documents does not establish that usable GIS boundaries, permit labels, licence rights or records matching the image dates are available. Before evaluating planning status, we will agree on the study city, date-matched records or boundaries, and labelling rules with our supervising faculty.

\begin{figure}[!htbp]\centering
\includegraphics[width=0.98\textwidth,height=0.70\textheight,keepaspectratio]{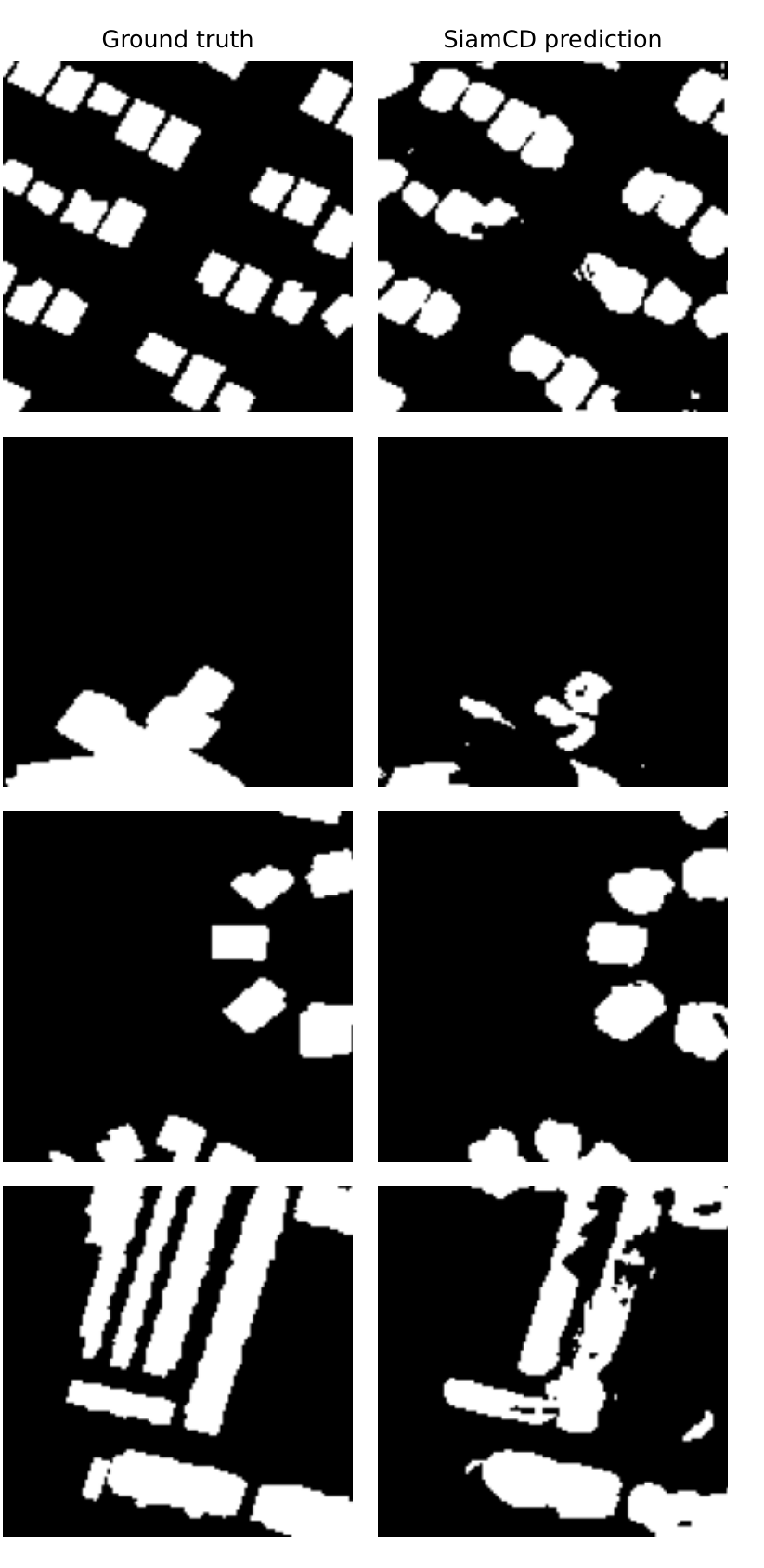}
\caption{Ground-truth change masks (left) and the reported SiamCD predicted masks (right) on LEVIR-CD test pairs, at threshold 0.5. Photographic panels are omitted. Examples are drawn by file index among patches with substantial ground-truth change, not hand-picked. Ground-truth annotations: Chen and Shi \cite{r5}, for academic use; predictions are from the model reported in Section 5.5.}\label{fig:11}
\end{figure}
\clearpage\section*{Reproducibility}
The available fixed-feature code and result files are in the public fork Harsh23Kashyap/Urban-Change-Detection, derived from Taranum01/Urban-Change-Detection. The repository includes the feature-based experiment scripts, src/multiseed\_eval.py and results/multiseed.json. Both datasets download automatically, and all runs used a CPU.

The per-seed values and ROC curves in Table 3 and Figures 3 and 4 come from a rerun of multiseed\_eval.py extended to save per-seed output. A separate script performed the 30-seed comparison using cached LEVIR-CD features. Figures 8 and 7 come from extra.py and calib.py using the same cached data.

The public fork at commit 17fce36 contains the experiment scripts, \path|paper_scripts/{seeds30,extra,calib}.py|, aggregate results and \path|paper_results/{seeds30,extra,calib}.json|. Model random states are fixed at zero, and split seeds are reported separately. Aggregation uses the population standard deviation.

The reconciliation rerun on 12 October 2026 used Python 3.10.12, NumPy 2.2.6 and scikit-learn 1.7.2. These versions describe this rerun only. A private archive retains the script, five per-split prediction arrays, environment record, metric checks and a SHA-256 manifest, without source imagery.

The code is MIT-licensed. That licence does not by itself give us the right to redistribute the source imagery, so those rights must be cleared before publication.

\end{document}